\def\year{2018}\relax
\documentclass[letterpaper]{article} %DO NOT CHANGE THIS
\usepackage{aaai18}  %Required
\usepackage{mydefs,amsmath}
\usepackage{times,graphicx}
\usepackage{latexsym}
\usepackage{url}
\usepackage{multirow}
\usepackage{times}  %Required
\usepackage{helvet}  %Required
\usepackage{courier}  %Required
\usepackage{xcolor}
\usepackage{url}  %Required
\usepackage{graphicx}  %Required
\usepackage{nameref}
\begin{document}
%\title{A Variational Autoencoder based Deep Generative Framework for Paraphrase Generation}
%\title{A Conditional Variational Autoencoder Framework for Paraphrase Generation}
\title{A Deep Generative Framework for Paraphrase Generation}

\author{
Ankush Gupta, Arvind Agarwal, Prawaan Singh, Piyush Rai\\
\{ankushgupta,arvagarw\}@in.ibm.com \and prawaan@iitk.ac.in \and piyush@cse.iitk.ac.in
}

% \author{AAAI Press\\
% Association for the Advancement of Artificial Intelligence\\
% 2275 East Bayshore Road, Suite 160\\
% Palo Alto, California 94303\\
% }

\maketitle
\begin{abstract}
Paraphrase generation is an important problem in NLP, especially in question answering, information retrieval, information extraction, conversation systems, to name a few. In this paper, we address the problem of generating paraphrases automatically. Our proposed method is based on a combination of deep generative models (VAE) with sequence-to-sequence models (LSTM) to generate paraphrases, given an input sentence. Traditional VAEs when combined with recurrent neural networks can generate free text but they are not suitable for paraphrase generation for a given sentence. We address this problem by conditioning the both, encoder and decoder sides of VAE, on the original sentence, so that it can generate the given sentence's paraphrases. Unlike most existing models, our model is simple, modular and can generate multiple paraphrases, for a given sentence. 
Quantitative evaluation of the proposed method on a benchmark paraphrase dataset demonstrates its efficacy, and its performance improvement over the state-of-the-art methods by a significant margin, whereas qualitative human evaluation indicate that the generated paraphrases are well-formed, grammatically correct, and are relevant to the input sentence. Furthermore, we evaluate our method on a newly released question paraphrase dataset, and establish a new baseline for future research.
%Experimental study of the proposed method on a benchmark paraphrase dataset demonstrates its efficacy, and its performance improvement over the state-of-the-art methods by a significant margin. Furthermore, we experiment on a newly released question paraphrase dataset and establish a new baseline for future research.
\end{abstract}

\section{Introduction}
\label{sec:intro}
Paraphrase generation is an important problem in many NLP applications such as question answering, information retrieval, information extraction, and summarization. 
%One of the prominent applications of paraphrase generation is in open question answering (QA) systems. 
QA systems are often susceptible to the way questions are asked; in fact, for knowledge-based (KB) QA systems, question paraphrasing is crucial for bridging the gap between questions asked by users and knowledge based assertions~\cite{fader2014open,yin2015answering}.
In an open QA system pipeline, question analysis and paraphrasing is a critical first step, in which a given question is reformulated by expanding it with its various paraphrases with the intention of improvement in recall, an important metric in the early stage of the pipeline. Similarly paraphrasing finds applications in information retrieval by generating query variants, and in machine translation or summarization by generating variants for automatic evaluation.   

In addition to being directly useful in QA systems, paraphrase generation is also important for generating training data for various learning tasks, such as question type classification, paraphrase detection, etc., that are useful in other applications. Question type classification has application in conversation systems, while paraphrase detection is an important problem for translation, summarization, social QA (finding closest question to FAQs/already asked question)~\cite{figueroa2013learning}. Due to the nature and complexity of the task, all of these problems suffer from lack of training data, a problem that can readily benefit from the paraphrase generation task.

Despite the importance of the paraphrase generation problem, there has been relatively little prior work in the literature, though much larger amount of work exists on paraphrase detection problem. Traditionally, paraphrase generation has been addressed using rule-based approaches~\cite{McKeown:1983:PQU:973241.973242,zhao2009application}, primarily due to the inherent difficulty of the underlying natural language generation problem. However, recent advances in deep learning, in particular generative models~\cite{bowman2015generating,chung2015recurrent}, have led to powerful, data-driven approaches to text generation.  

In this paper, we present a deep generative framework for automatically generating paraphrases, given a sentence. Our framework combines the power of sequence-to-sequence models, specifically the long short-term memory (LSTM)~\cite{hochreiter1997long}, and deep generative models, specifically the variational autoencoder (VAE)~\cite{kingma2013auto,rezende2014stochastic}, to develop a novel, end-to-end deep learning architecture for the task of paraphrase generation.

In contrast to the recent usage of VAE for sentence generation~\cite{bowman2015generating}, a key differentiating aspect of our proposed VAE based architecture is that it needs to generate paraphrases, \emph{given} an original sentence as input. That is, the generated paraphrased version of the sentence should capture the essence of the original sentence. Therefore, unconditional sentence generation models, such as~\cite{bowman2015generating}, are not suited for this task. To address this limitation, we present a mechanism to \emph{condition} our VAE model on the original sentence for which we wish to generate the paraphrases. In the past, conditional generative models~\cite{sohn2015learning,kingma2014semi} have been applied in computer vision to generate images conditioned on the given class label. Unlike these methods where number of classes are finite, and do not require any intermediate representation, our method conditions both the sides (i.e. encoder and decoder) of VAE on the intermediate representation of the input question obtained through LSTM.

One potential approach to solve the paraphrase generation problem could be to use existing sequence-to-sequence models~\cite{sutskever2014sequence}, in fact, one variation of sequence-to-sequence model using stacked residual LSTM~\cite{prakash2016neural} is the current state of the art for this task. However, most of the existing models for this task including stacked residual LSTM, despite having sophisticated model architectures, lack a principled generative framework. In contrast, our deep generative model enjoys a simple, modular architecture, and can generate not just a single but \emph{multiple}, semantically sensible, paraphrases for any given sentence.

It is worth noting that existing models such as sequence-to-sequence models, when applied using beam search, are not able to produce multiple paraphrases in a principled way. Although one can choose top $k$ variations from the ranked results returned by beam-search, $k_{th}$ variation will be qualitatively worse (by the nature of beam-search) than the first variation. This is in contrast to the proposed method where all variations will be of relatively better quality since they are the \textit{top} beam-search result, generated based on different $z$ sampled from a latent space. We compare our framework with various sophisticated sequence-to-sequence models including the state-of-the-art stacked residual model~\cite{prakash2016neural} for paraphrase generation, and show its efficacy on benchmark datasets, on which it outperforms the state-of-the-art by significant margins. Due to the importance of the paraphrase generation task in QA system, we perform a  comprehensive evaluation of our proposed model on the recently released Quora questions dataset\footnote{https://data.quora.com/First-Quora-Dataset-Release-Question-Pairs}, and demonstrates its effectiveness for the task of question paraphrase generation through both quantitative metrics, as well as qualitative analysis. Human evaluation indicate that the paraphrases generated by our system are well-formed, and grammatically correct for the most part, and are able to capture new concepts related to the input sentence.

%Our contribution in this paper is as follows:
%We present a supervised VAE based method for generating paraphrase automatically. Our method can generate multiple paraphrases of a given question. We experiment with the question paraphrase data and show that out method generates surprisingly well paraphrases. Since the method is equally applicable to general paraphrase generation, we apply it to a benchmark dataset and show that our method outperform the much sophisticated state-of-the-art method by a significant margin. 

\section{Methodology}
\label{sec:approach}
Our framework uses a variational autoencoder (VAE) as a generative model for paraphrase generation. In contrast to the standard VAE, however, we additionally condition the encoder and decoder modules of the VAE on the \emph{original} sentence. This enables us to generate paraphrase(s) specific to an input sentence at test time. In this section, we first provide a brief overview of VAE, and then describe our framework.

\begin{figure}[!htbp]
	\begin{center}
     	\includegraphics[scale=0.3]{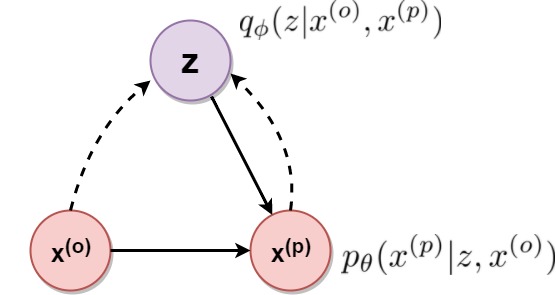}
    \end{center}
    \vspace{-0.2in}
    \caption{\small A macro-view of our model: the paraphrase generation model is also conditioned on the original sentence}
    \label{fig:macro}
\end{figure}

\begin{figure*}[!htbp]
	\begin{center}
    	\includegraphics[scale=0.14]{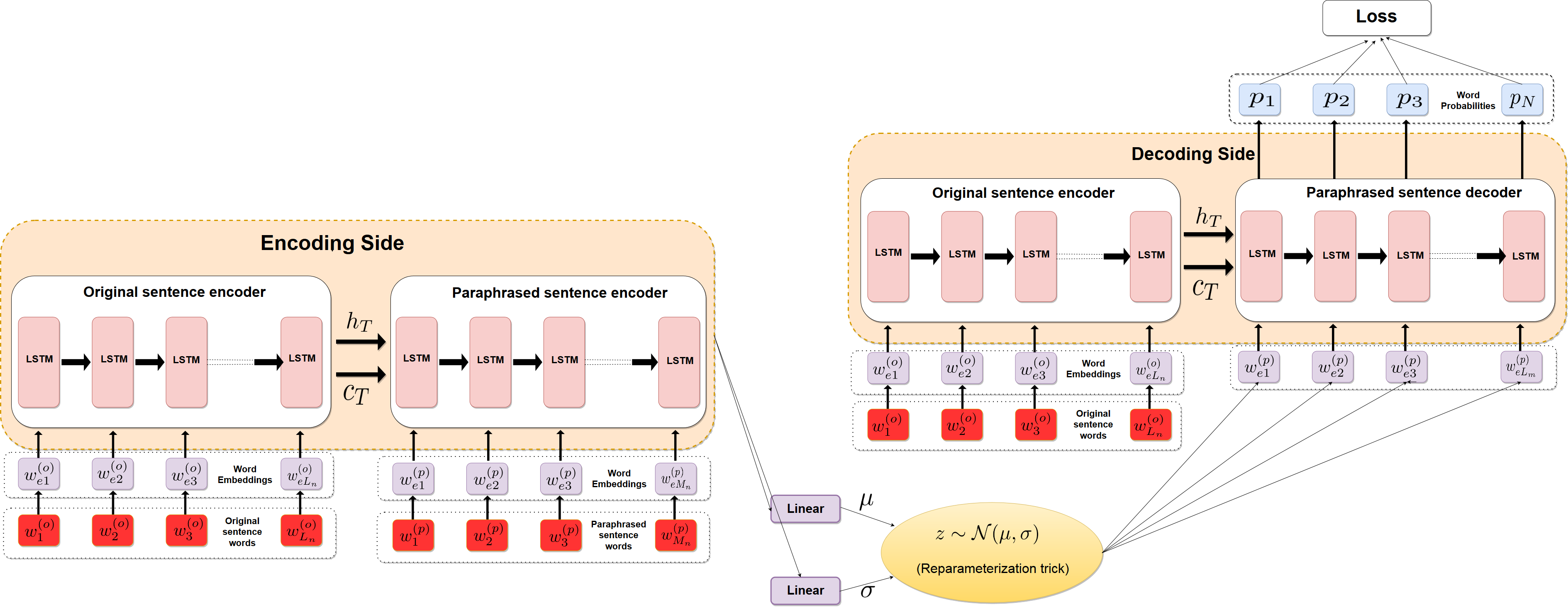}
    \end{center}
    \vspace{-0.2in}
    \caption{\small The block diagram of our VAE-LSTM architecture for paraphrase generation}
    \label{fig:block}
\end{figure*}

\subsection{Variational Autoencoder (VAE)}

The VAE~\cite{kingma2014auto,rezende2014stochastic} is a deep generative latent variable model that allows learning rich, nonlinear representations for high-dimensional inputs. The VAE does so by learning a latent representation or ``code'' $\zv \in \mathbb{R}^K$ for an input $\xv \in \mathbb{R}^D$ such that the original input $\xv$ can be well reconstructed from the latent code $\zv$. In contrast to the standard autoencoder~\cite{goodfellow2016deep} which learns, for any input $\xv$, a \emph{deterministic} latent code $\zv$ via a deterministic encoder function $q_\phi$, the VAE encoder is actually a \emph{posterior distribution} $q_{\phi}(\zv|\xv)$ (also known as the \emph{recognition model}) over the latent code $\zv$. The posterior $q_{\phi}(\zv|\xv)$ is usually assumed to be a Gaussian distribution $\Ncal(\mu(\xv),\text{diag}(\sigma^2(\xv)))$, and the parameters $\phi = \{\mu(\xv),\sigma^2(\xv)\}$ are nonlinear transformations of the input $\xv$ and are the outputs of feedforward neural networks that take $\xv$ as input. The VAE also encourages its posterior distribution $q_{\phi}(\zv|\xv)$ to be close to the prior $p(\zv)$, which is typically taken as a standard normal distribution $\Ncal(0,\Imat)$.

The VAE also consists of a \emph{decoder} model, which is another distribution $p_\theta(\xv|\zv)$ that takes as input a \emph{random} latent code $\zv$ and produces an observation $\xv$. The parameters of the decoder distribution $\theta$ are defined by the outputs of another feedforward neural networks, akin to the VAE encoder model. 

The parameters defining the VAE are learned by maximizing the following objective:
\begin{multline} 
  \Lcal(\theta,\phi;\xv) = \mathbb{E}_{q_\phi(\zv|\xv)}[\log p_\theta(\xv|\zv)] \\ - KL(q_\phi(\zv|\xv)||p(\zv))
  \label{eq:elbo}
\end{multline}
Here KL stands for the KL divergence. Eq.~\ref{eq:elbo} provides a lower bound on the model evidence $p(\xv|\theta,\phi)$ and the VAE parameters are learned by maximizing this lower bound~\cite{kingma2014auto,rezende2014stochastic}.

Endowing the latent code $\zv$ with a distribution ``prepares'' the VAE decoder for producing realistic looking inputs even when $\zv$ is a \emph{random} latent code not representing the encoding of any of the previously seen inputs. This makes VAE very attractive for generative models for complex data, such as images and text data such as sentences. 

In particular,~\cite{bowman2015generating} presented a text-generation model in which the encoder and decoder were modeled by long short-term memory (LSTM) networks. Moreover, training tricks such as KL-term annealing and dropout of inputs of the decoder were employed to circumvent the problems encountered when using the standard VAE for the task of modeling text data. Our work is in a similar vein but the key difference lies in the design of a novel VAE-LSTM architecture, specifically customized for the paraphrase generation task, where the training examples are given in form of pairs of sentences (original sentence and its paraphrased version), and both encoder and decoder of the VAE-LSTM are conditioned on the original sentence. We describe our VAE-LSTM architecture in more detail in the next section.

\subsection{Model Architecture}
\label{sec:modelarch}
Our training data is provided in form of $N$ pairs $\{\sv^{(o)}_n,\sv^{(p)}_n\}_{n=1}^N$, with each pair consisting of the \emph{original} sentence (denoted by superscript $o$) and its \emph{paraphrase} (denoted by superscript $p$). For the $n^{th}$ pair, $\sv^{(o)}_n = \{w_1^{(o)},\ldots,w_{L_n}^{(o)}\}$ and $\sv^{(p)}_n = \{w_1^{(p)},\ldots,w_{M_n}^{(p)}\}$ denote the set of $L_n$ words from the \emph{original} sentence and $M_n$ words from its \emph{paraphrase}, respectively. In the following description, we will omit explicitly using the pair index $n$; e.g., we will denote a pair of original sentence and its paraphrase simply by $\sv^{(o)}$ and $\sv^{(p)}$, respectively. We will also use $\xv^{(o)}$ and $\xv^{(p)}$ to denote the vector space representations of the original sentence and its paraphrase, respectively. These representations will be learned using LSTM networks, whose parameters will be learned in an end-to-end fashion, with the rest of the model.

Fig.~\ref{fig:macro} shows a macro view (without the LSTM) of our proposed model architecture, which is essentially a VAE based generative model for each paraphrase's vector representation $\xv^{(p)}$, which in turn is generated by a latent code $\zv$ and the original sentence $\xv^{o}$. In addition, unlike the standard VAE, note that our VAE decoder model $p_{\theta}(\xv^{(p)}|\zv,\xv^{(o)})$ is also \emph{conditioned} on the vector representation $\xv^{(o)}$ of the original sentence. In particular, as Fig.~\ref{fig:macro} shows, the VAE encoder as well as decoder are conditioned on the original sentence.

A detailed zoomed-in view of our model architecture is shown in Fig.~\ref{fig:block}, where we show all the components, including the LSTM encoders and decoders. In particular, our model consists of three LSTM encoders and one LSTM decoder (thus a total of four LSTMs), which are employed by our VAE based architecture as follows:
\begin{itemize}
	\item \textbf{VAE Input (Encoder) Side:} As shown in Fig.~\ref{fig:block}, two of the LSTM encoders are used on the VAE's input side. The first one converts the original sentence $\sv^{(o)}$ into its vector representation $\xv^{(o)}$, which is fed, along with the paraphrase version $\sv^{(p)}$ of this sentence, to the next LSTM encoder. The output of this LSTM encoder ($\xv^{(p)}$) is passed through a feedforward neural network to produce the mean and variance parameters i.e., $\phi$, of the VAE encoder.
    \item \textbf{VAE Output (Decoder) Side:} As shown in Fig.~\ref{fig:block}, the VAE's output side uses an LSTM decoder which takes as input (1) the latent code $\zv$,  and (2) vector representation $\xv^{(o)}$ (produced by the third LSTM encoder) of the original sentence. The vector representation $\xv^o$ is used to initialize the LSTM decoder by feeding it to the first stage of the decoder, in contrast to the latent code $\zv$ which is fed to each stage of the LSTM decoder (after being concatenated with the output of previous LSTM stage). Thus both $\zv$ and $\xv^o$ are used to reconstruct the paraphrased sentence $\sv^{(p)}$.
\end{itemize}
Similar to the VAE, the variational lower-bound of the proposed model is given by:
\small 
\begin{multline} 
  \Lcal(\theta,\phi;\xv^{(p)},\xv^{(o)}) = \mathbb{E}_{q_\phi(\zv|\xv^{(o)},\xv^{(p)})}[\log p_\theta(\xv^{(p)}|\zv,\xv^{(o)})] \\ - KL(q_\phi(\zv|\xv^{(o)},\xv^{(p)})||p(\zv))
  \label{eq:elbo2}
\end{multline}
\normalsize
Maximizing the above lower bound trades off the expected reconstruction of the paraphrased sentence's representation $\xv^{(p)}$ (given $\xv^{(o)}$), while ensuring that the posterior of $\zv$ is close to the prior. We train our model following the same training procedure as employed in~\cite{bowman2015generating}.
 
\section{Related Work}

%In addition to the prior work described in the Introduction section, here we describe some other prior work on paraphrase generation. 
% TBD: Talk about GAN?

%\textbf{Paraphrase Generation:} 
The task of generating paraphrases of a given sentence has been dealt in great depth and in various different types of approaches. Work has been done to use Statistical Machine Translation based models for generating paraphrases. ~\cite{quirk2004monolingual} apply SMT tools, trained on large volumes of sentence pairs from news articles. ~\cite{zhao2008combining} proposed a model that uses multiple resources to improve SMT based paraphrasing, paraphrase table and feature function which are then combined in a log-linear SMT model. Some old methods use data-driven methods and hard coded rules such as ~\cite{madnani2010generating}, ~\cite{mckeown1983paraphrasing}. ~\cite{hassan2007unt} proposes a system for lexical substitution using thesaurus methods. ~\cite{kozlowski2003generation} pairs elementary semantic structures with their syntactic realization and generate paraphrases from predicate/argument structure. As mentioned in ~\cite{prakash2016neural}, the application of deep learning models to paraphrase generation has not been explored rigorously yet. This is one of the first major works that used deep architecture for paraphrase generation and introduce the residual recurrent neural networks.

Finally, our work is also similar in spirit to other generative models for text, e.g. controllable text generation~\cite{hu2017controllable}, which combines VAE and explicit constraints on independent attribute controls. Other prior works on VAE for text generation include~\cite{bowman2015generating,semeniuta2017hybrid} which used VAEs to model holistic properties of sentences such as style, topic and various other syntactic features.

\section{Experiments}
\label{sec:exp}
% \begin{table}[]
% \centering
% \caption{Dataset Details}
% \label{tab:dataset-details}
% \begin{tabular}{|l|l|l|p{1.5cm}|}
% \hline
% \textbf{Dataset} & \textbf{Training} & \textbf{Test}   & \textbf{Vocab Size} \\ \hline
% Quora   &  50K/100K/150K        & 4K  &                 \\ \hline
% MSCOCO  &   168K       & 40K &         \\       \hline
% \end{tabular}
% \end{table}

\begin{table}[tb]
\vspace{-2em}
\centering
\small
\caption{\scriptsize Different models compared in the evaluation study.}
\label{tab:models}
\begin{tabular}{|l|l|}
\hline
\textbf{Models} & \textbf{Reference} \\ \hline
Seq-to-Seq   &    \cite{sutskever2014sequence}   \\ \hline
With Attention  &   \cite{bahdanau2014neural}    \\       \hline
Bi-directional LSTM  &    \cite{graves2013hybrid}   \\       \hline
Residual LSTM  &   \cite{prakash2016neural}   \\       \hline
Unsupervised & Ours but baseline \\ \hline
VAE-S & Ours but baseline \\ \hline
VAE-SVG & Ours \\ \hline
VAE-SVG-eq & Ours \\ \hline
\end{tabular}
\end{table}

In this section, we describe the datasets, experimental setup, evaluation metrics and the results of our experiments.
\subsection{Datasets}
\label{sec:datasets}
We evaluate our framework on two datasets, one of which (MSCOCO) is for the task of standard paraphrase generation and the other (Quora) is a newer dataset for the specific problem of question paraphrase generation.

\textbf{MSCOCO}~\cite{lin2014microsoft}: This dataset, also used previously to evaluate paraphrase generation methods~\cite{prakash2016neural}, contains human annotated captions of over 120K images. Each image contains five captions from five different annotators. This dataset is a standard benchmark dataset for image caption generation task. In majority of the cases, annotators describe the most prominent object/action in an image, which makes this dataset suitable for the paraphrase generation task.  The dataset has separate division for training and validation. \textit{Train 2014} contains over 82K images and \textit{Val 2014} contains over 40K images. From the five captions accompanying each image, we randomly omit one caption, and use the other four as training instances (by creating two source-reference pairs). Because of the free form nature of the caption generation task~\cite{vinyals2015show}, some captions were very long. We reduced those captions to the size of 15 words (by removing the words beyond the first 15) in order to reduce the training complexity of the models, and also to compare our results with previous work~\cite{prakash2016neural}. 
Some examples of input sentence and their generated paraphrases can be found in Table~\ref{tab:good-examples-mscoco}.

\textbf{Quora}: Quora released a new dataset in January 2017. The dataset consists of over 400K lines of potential question duplicate pairs. Each line contains IDs for each question in the pair, the full text for each question, and a binary value that indicates whether the questions in the pair are truly a duplicate of each-other.\footnote{https://data.quora.com/First-Quora-Dataset-Release-Question-Pairs}. Wherever the binary value is 1, the question in the pair are not identical; they are rather paraphrases of each-other. So, for our study, we choose all such question pairs with binary value 1. There are a total of $155K$ such questions. In our experiments, we evaluate our model on 50K, 100K and 150K training dataset sizes. For testing, we use 4K pairs of paraphrases. Some examples of question and their generated paraphrases can be found in Table~\ref{tab:good-examples-quora}.\\

\subsection{Baselines}
\label{subsec:baselines}
We consider several state-of-the-art baselines for our experiments. These are described in Table~\ref{tab:models}. For MSCOCO, we report results from four baselines, with the most  important of them being by~\cite{prakash2016neural} using residual LSTM. Residual LSTM is also the current state-of-the-art on the MSCOCO dataset. For the Quora dataset, there were no known baseline results, so we compare our model with (1) standard VAE model i.e., the unsupervised version, and (2) a ``supervised'' variant VAE-S of the unsupervised model. In the unsupervised version, the VAE generator reconstructs multiple variants of the input sentence using the VAE generative model trained only using the original sentence (without their paraphrases); in VAE-S, the VAE generator {\it generates} the paraphrase conditioned on the original sentence, just like in the proposed model. This VAE-S model can be thought of as a variation of the proposed model where we remove the encoder LSTM related to the paraphrase sentence from the encoder side. Alternatively, it is akin to a variation of VAE where decoder is made supervised by making it to generate ``paraphrases" (instead of the reconstructing original sentence as in VAE) by conditioning the decoder on the input sentence.

% \usepackage{multirow}
\begin{table*}[htb]
\centering
\scriptsize
\caption{\scriptsize Results on MSCOCO dataset. Higher BLEU and METEOR score is better, whereas lower TER score is better. "Measure" column denotes the way metrics are computed over multiple paraphrases.}
\label{tab:mscoco-results}
\begin{tabular}{|c|l|l|l|l|l|l|}
\hline
\multicolumn{1}{|l|}{}                                 &             & \multicolumn{5}{c|}{\textbf{MSCOCO}}                                                                                          \\ \hline
\textbf{Model}                                       & Measure     & Beam size & {\tiny Layers} & {\tiny BLEU}  & {\tiny METEOR} & {\tiny TER}   \\ \hline
\textbf{Seq-to-Seq \cite{sutskever2014sequence}}   & -           & 10                                                  & 2                & 16.5            & 15.4             & 67.1            \\ \hline
\textbf{With Attention \cite{bahdanau2014neural}}  & -           & 10                                                  & 2                & 18.6            & 16.8             & 63.0            \\ \hline
\textbf{Seq-toSeq \cite{sutskever2014sequence}}    & -           & 10                                                  & 4                & 28.9            & 23.2             & 56.3            \\ \hline
\textbf{Bi-directional \cite{graves2013hybrid}}    & -           & 10                                                  & 4                & 32.8            & 24.9             & 53.7            \\ \hline
\textbf{With Attention \cite{bahdanau2014neural} } & -           & 10                                                  & 4                & 33.4            & 25.2             & 53.8            \\ \hline
\textbf{Residual LSTM \cite{prakash2016neural}}    & -           & 10                                                  & 4                & 37.0            & 27.0             & 51.6            \\ \hline
\textbf{Unsupervised}                                & -           & No beam                                                   & 1, 2             & 12.8            & 17.5             & 78.8            \\ \hline
\multirow{3}{*}{\textbf{VAE-S}}                                 & Avg         & No beam                                             & 1, 2             & 7.0             & 14.0             & 82.3            \\ \cline{2-7} 
                                                       & best BLEU   & No beam                                                   & 1, 2             & 11.3            & 16.8             & 76.5            \\ \cline{2-7} 
                                                       & best METEOR & No beam                                                   & 1, 2             & 11.0            & 17.7             & 78.8            \\ \hline
\multirow{6}{*}{\textbf{VAE-SVG (our)}}                         & Avg         & No beam                                                   & 1, 2              & 39.2 & 29.2  & 43.6 \\ \cline{2-7} 
                                                       & best BLEU   & No beam                                              & 1, 2              & 41.1 & 30.3  & 42.3 \\ \cline{2-7} 
                                                       & best METEOR & No beam                                             & 1, 2             & \textbf{41.7} & 30.8  & 41.7 \\ \cline{2-7} 
                                                       & Avg         & 10                                                  & 1, 2             & 41.3            & 30.9             & \textbf{40.8}            \\ \cline{2-7} 
                                                       & best BLEU   & 10                                                  & 1, 2             &     40.9            &   30.7               &   42.0              \\ \cline{2-7} 
                                                       & best METEOR & 10                                                  & 1, 2             &      41.3           &   \textbf{31.0}               &   41.6              \\ \hline
\multirow{6}{*}{\textbf{VAE-SVG-eq (our)}}                      & Avg         & No beam                                             & 1, 2             & 37.3            & 28.5             & 45.1            \\ \cline{2-7} 
                                                       & best BLEU   & No beam                                             & 1, 2             & 39.2            & 29.5             & 43.9            \\ \cline{2-7} 
                                                       & best METEOR & No beam                                             & 1, 2             & 39.8            & 30.0             & 43.4            \\ \cline{2-7} 
                                                       & Avg         & 10                                                  &  1, 2                & 39.6            & 30.2             & 42.3            \\ \cline{2-7} 
                                                       & best BLEU   & 10                                                  &  1, 2                &    39.3             &   30.1               &    43.5             \\ \cline{2-7} 
                                                       & best METEOR & 10                                                  & 1, 2                 &   39.7              &    30.4              &   43.2              \\ \hline
\end{tabular}
\end{table*}

\begin{table*}[htb]
\centering
\scriptsize
\caption{\scriptsize Results on Quora dataset. Higher BLEU and METEOR score is better, whereas lower TER score is better.}
\label{tab:quora-results}
\begin{tabular}{|c|l|l|l|l|l|l|l|l|l|l|}
\hline
\multicolumn{1}{|l|}{}                 &             & \multicolumn{9}{c|}{\textbf{Quora}}                                                                                                                                \\ \hline
\multicolumn{1}{|l|}{}                 &             & \multicolumn{3}{c|}{50K}                             & \multicolumn{3}{c|}{100K}                            & \multicolumn{3}{c|}{150K}                            \\ \hline
\textbf{Model}                       & Measure     & {\tiny BLEU}  & {\tiny METEOR} & {\tiny TER}   & {\tiny BLEU}  & {\tiny METEOR} & {\tiny TER}   & {\tiny BLEU}  & {\tiny METEOR} & {\tiny TER}   \\ \hline
\textbf{Unsupervised  (baseline)}                 & -           & 8.3             & 12.2             & 83.7            & 10.6            & 14.3             & 79.9            & 11.4            & 14.5             & 78.0            \\ \hline
\multirow{3}{*}{\textbf{\begin{tabular}[c]{@{}c@{}}VAE-S\\ (baseline)\end{tabular}}}      & Avg         & 11.9            & 17.4             & 77.7            & 13.0            & 18.4             & 76.8            & 14.2            & 19.0             & 74.8            \\ \cline{2-11} 
                                       & best BLEU   & 15.8            & 20.1             & 69.4            & 17.5            & 21.6             & 67.1            & 19.8            & 22.6             & 63.9            \\ \cline{2-11} 
                                       & best METEOR & 15.6            & 21.1             & 71.5            & 17.5            & 22.7             & 69.5            & 19.7            & 23.8             & 66.9            \\ \hline
\multirow{3}{*}{\textbf{\begin{tabular}[c]{@{}c@{}}VAE-SVG \\ (ours)\end{tabular}}}    & Avg         & 13.8            & 18.7             & 68.2            & 18.6            & 21.9             & 60.6            & 25.0            & 25.1             & 52.5            \\ \cline{2-11} 
                                       & best BLEU   & 17.1            & 21.3             & 63.1            & 22.5            & 24.6             & 55.7            & 30.3            & 28.5             & 47.3            \\ \cline{2-11} 
                                       & best METEOR & 17.1            & \textbf{22.2}  & 63.8            & 22.4            & \textbf{25.5}  & 55.6            & 30.3            & 29.2             & 47.1            \\ \hline
\multirow{6}{*}{\textbf{\begin{tabular}[c]{@{}c@{}}VAE-SVG-eq\\ (ours)\end{tabular}}} & Avg         & 13.9            & 18.8             & 67.1            & 19.0            & 21.7             & 60.0            & 26.2            & 25.7             & 52.1            \\ \cline{2-11} 
                                       & best BLEU   & \textbf{17.4} & 21.4             & \textbf{61.9} & \textbf{22.9} & 24.7             & 55.0            & 31.4            & 29.0             & 46.8            \\ \cline{2-11} 
                                       & best METEOR & 17.3            & \textbf{22.2}  & 62.6            & \textbf{22.9} & \textbf{25.5}  & \textbf{54.9} & 32.0 & 30.0  & 46.1 \\ \cline{2-11}
       								   & Avg (beam=10)        & -    & -     & - & - & - & - & 37.1 & 32.0 & 40.8  \\ \cline{2-11} 
                                       & best BLEU (beam=10)  & - 	& -     & - & - & - & - & 38.0 & 32.9 & 40.0  \\ \cline{2-11} 
                                       & best METEOR (beam=10) & -    & -  	& - & - & - & - & \textbf{38.3} & \textbf{33.6} & \textbf{39.5} \\ \hline
\end{tabular}
\end{table*}

\subsection{Experimental Setup}
\label{subsec:training}
Our framework primarily uses the following experimental setup. These settings are directly borrowed from an existing implementation\footnote{\url{https://github.com/kefirski/pytorch_RVAE}} of the paper~\cite{bowman2015generating}, and were not fine tuned for any of the datasets. In our setup, we do not use any external word embeddings such as Glove; rather we train these as part of the model-training. The dimension of the embedding vector is set to 300, the dimension of both encoder and decoder is 600, and the latent space dimension is 1100. The number of layers in the encoder is 1 and in decoder 2. Models are trained with stochastic gradient descent with learning rate fixed at a value of $5\times10^{-5}$ with dropout rate of $30\%$. Batch size is kept at 32. Models are trained for a predefined number of iterations, rather than a fixed number of epochs. In each iteration, we sequentially pick the next batch. A fixed number of iterations makes sure that we do not increase the training time with the amount of data. When the amount of data is increased, we run fewer passes over the data as opposed to the case when there is less data. Number of units in LSTM are set to be the maximum length of the sequence in the training data. 
%The training time of Quora data set for supervised model was approximately 20 hours while for MSCOCOC dataset 11 hours. The test time for Quora was 30 min and for MSCOCO 6 hours. 

%############################################################################
\subsection{Evaluation}
\label{sec:evaluation}

\paragraph{Quantitative Evaluation Metrics}
For quantitative evaluation, we use the well-known automatic evaluation metrics\footnote{We used the software available at https://github.com/jhclark/multeval} in machine translation domain : BLEU~\cite{papineni2002BLEU}, METEOR~\cite{lavie2007meteor}, and Translation Error Rate (TER)~\cite{snover2006study}. Previous work has shown that these metrics can perform well
for the paraphrase recognition task~\cite{madnani2012re} and correlate well with human judgments in evaluating generated paraphrases~\cite{wubben2010paraphrase}.
BLEU considers exact match between reference paraphrase(s) and system generated paraphrase(s) using the concept of modified n-gram precision and brevity penalty. METEOR also uses stemming and synonyms (using WordNet) while calculating the score and is based on a combination of unigram-precision and unigram-recall with the reference paraphrase(s). TER is based on the number of edits (insertions, deletions, substitutions, shifts) required for a human to convert the system output into one of the reference paraphrases.

\paragraph{Qualitative Evaluation Metrics}
To quantify the aspects that are not addressed by automatic evaluation metrics, human evaluation becomes necessary for our problem. We collect human judgments on 100 random input sentences from both MSCOCO and Quora dataset. Two aspects are verified in human evaluation : \textit{Relevance} of generated paraphrase with the input sentence and \textit{Readability} of generated paraphrase. Six Human evaluators (3 for each dataset) assign a score on a continuous scale of 1-5 for each aspect per generated paraphrase, where 1 is worse and 5 is best.

%We adopt the following definition :

%\textbf{Relevance} : How relevant is the generated paraphrase to the given input sentence/question? \\
%1	Completely Irrelevant \\
%2	Irrelevant with few Relevant words \\
%3	Some what Relevant \\
%4	Relevant but with some Irrelevant words \\
%5	Very Relevant \\

%\textbf{Readability} : How grammatically correct is the generated paraphrase? \\
%1	Does not make sense at all \\
%2	Took some efforts to make sense \\
%3	Grammatically incorrect but makes sense \\
%4	Grammatical correct with some additional words, readable \\
%5	Absolutely grammatically correct \\

\subsection{Model Variations}
In addition to the model proposed in~\nameref{sec:approach} Section, we also experiment with another variation of this model. In this variation, we make the encoder of original sentence same on both sides i.e.  encoder side and the decoder side. We call this model VAE-SVG-eq (SVG stands for sentence variant generation). The motivation for this variation is that having same encoder reduces the number of model parameters, and hopefully helps in learning.

\newcommand{\specialcell}[2][c]{%
  \begin{tabular}[#1]{@{}l@{}l@{}}#2\end{tabular}}

\vspace{-0.2in}
{\small
\begin{table}[tbh]
\scriptsize
\centering
\caption{\scriptsize Some examples of paraphrases generated on MSCOCO Dataset.}
\label{tab:good-examples-mscoco}
\begin{tabular}{|l|p{2.6in}|}
%\hline
%          & \textbf{MSCOCO}                                          \\ 
\hline
Source    & A man with luggage on wheels standing next to a white van .              \\ \hline
Reference & A white Van is {\color{red}{driving}} through a {\color{red}{busy street}} .         \\ \hline
Generated & \specialcell[]{
A {\color{blue}{young man}} standing {\color{blue}{in front of an airport}} .\\
A {\color{blue}{yellow van}} is {\color{blue}{parked}} at the {\color{blue}{busy street}} .\\
A white van is in the {\color{blue}{middle of a park}} .
}       \\ 
\hline          
\hline
Source    & A table full of vegetables and fruits piled on top of each other .                        \\ \hline
Reference & \textcolor{red}{Group of mixed} vegetables sitting on a {\color{red}{counter top}} in a kitchen .          \\ \hline
Generated & 
\specialcell[]{{\color{blue}{Several plates of fruits}} sitting on a {\color{blue}{table top}}  in a kitchen \\
{\color{blue}{Assortment of fruits and vegetables}}  on a {\color{blue}{wooden table}}  in a kitchen . \\
{\color{blue}{Several types of fruit}} sitting on a {\color{blue}{counter top}} in a kitchen .}   \\ \hline
\hline
Source    & Large motorcycle sitting on a grassy area in a line .    \\ \hline 
Reference & {\color{red}{Older}} motorcycle {\color{red}{displayed}} on grass along with several old cars .  \\ \hline
Generated & \specialcell[]{{\color{blue}{Black motorcycle parked}} on the {\color{blue}{roadside}} next to a house .\\
{\color{blue}{Black motorcycle rider}} on dirt next to a group of people .\\
{\color{blue}{Green motorcycle parked on display}} outside with several old {\color{blue}{buses}} .}         \\ \hline
\end{tabular}
\vspace{-0.1in}
\end{table}
}

\begin{table}[tbh]
\centering
\scriptsize
\caption{\scriptsize Some examples of question paraphrases generated on Quora Dataset.}
\label{tab:good-examples-quora}
\begin{tabular}{|l|p{2.6in}|}
\hline
%          & \textbf{Quora}                                          \\ \hline
Source    & What is my old Gmail account ?                        \\ \hline
Reference & How can you {\color{red}{find all of your}} Gmail accounts ?         \\ \hline
Generated & \specialcell[]{Is there any way to {\color{blue}{recover}} my Gmail account ?\\How can I find my old Gmail {\color{blue}{account number}}?\\How can I {\color{blue}{get}} the old Gmail {\color{blue}{account password}} ?}  \\ \hline
\hline
Source    & Which is the best training institute in Pune for digital marketing and why ?                        \\ \hline
Reference & Which is the best {\color{red}{digital marketing}} training institute in Pune ?  \\ \hline
Generated & \specialcell[]{Which is the best institute for digital training in Pune ?\\Which is the best digital {\color{blue}{Tech training}} institute in {\color{blue}{Hyderabad}} ?\\Which is the best digital marketing training {\color{blue}{center}} in Pune ?}         \\ \hline
\hline
Source    & What are my options to making money online ?              \\ \hline
Reference & How can we {\color{red}{earn}} money through online ?         \\ \hline
Generated & \specialcell[]{How can I {\color{blue}{make}} money online ?\\What are {\color{blue}{ways}} of earning money online ?\\How can I {\color{blue}{profitable}} earn money online ?}       \\ \hline

\end{tabular}
\vspace{-2em}
\end{table}

\subsection{Results}
\label{sec:results}
We perform experiments on the above mentioned datasets, and report, both qualitative and quantitative results of our approach. The qualitative results for MSCOCO and Quora datasets are given in Tables~\ref{tab:good-examples-mscoco} and~\ref{tab:good-examples-quora} respectively. In these tables, Red and Blue colors denote interesting phrases which are different in the ground truth and generated variations respectively w.r.t. the input sentence. From both the tables, we see that variations contain many interesting phrases such as  \textit{in front of an airport}, \textit{busy street}, \textit{wooden table}, \textit{recover}, \textit{Tech training} etc. which were not encountered in input sentences. Furthermore, the paraphrases generated by our system are well-formed, semantically sensible, and grammatically correct for the most part. For example, for the MSCOCO dataset, for the input sentence 
{\it A man with luggage on wheels standing next to a white van.}, one of the variants {\it A young man standing in front of an airport.} is able to figure out that the situation pertains to ``waiting in front of an airport", probably from the phrases \textit{standing} and \textit{luggage on wheels}. Similarly, for the Quora dataset, for the question {\it What is my old Gmail account?}, one of the variants is {\it Is there any way to recover my Gmail account?} which is very similar --but not the same-- to the paraphrase available in the ground truth. It is further able to figure out that the input sentence is talking about \textit{recovering} the account. Another variant {\it How can I get the old Gmail account password?} tells us that accounts are related to the password, and recovering the account might mean recovering the password as well.

% We perform experiments on the above mentioned datasets, and report, both qualitative and quantitative results of our approach. The qualitative results for MSCOCO and Quora datasets are given in Tables~\ref{tab:good-examples-mscoco} and~\ref{tab:good-examples-quora} respectively. As we can see from the tables, that paraphrases generated by our system are well-formed, semantically sensible, and grammatically correct for the most part. For example, for the MSCOCO dataset, for the input sentence {\it A bowl of soup that has some carrots , shrimp , and noodles in it .}, one of the variants {\it The dinner dish is in the bowl that is for served .} is able to figure out that it is dinner and is served. Similarly, for the Quora dataset, for the question {\it What is my old Gmail account ?}, one of the variants is {\it How can I find all my Gmail account?} which is very similar --but not the same-- to the variant available in ground truth. Another variant {\it How can I get the old Gmail account password ?} tells us that accounts are related to the password, and recovering the account might mean recovering the password as well.

In Tables~\ref{tab:mscoco-results} and~\ref{tab:quora-results}, we report the quantitative results from various models for the MSCOCO and Quora datasets respectively.  Since our models generate multiple variants of the input sentence, one can compute multiple metrics with respect to each of the variants. In our tables, we report average and best of these metrics. For average, we compute the metric between each of the generated variants and the \textit{ground truth}, and then take the average. For computing the best variant, while one can use the same strategy, that is, compute the metric between each of the generated variants and the \textit{ground truth}, and instead of taking average find the best value but that would be unfair. Note that in this case, we would be using the ground truth to compute the best which is not available at test time. Since we cannot use the ground truth to find the best value, we instead use the metric between the {\it input sentence} and the variant to get the best variant, and then report the metric between the best variant and the \textit{ground truth}. Those numbers are reported in the {\it Measure} column with row best-BLEU/best-METEOR. 
\vspace{-0.1in}
\begin{figure}[!htbp]
	\begin{center}
    	\includegraphics[scale=0.35]{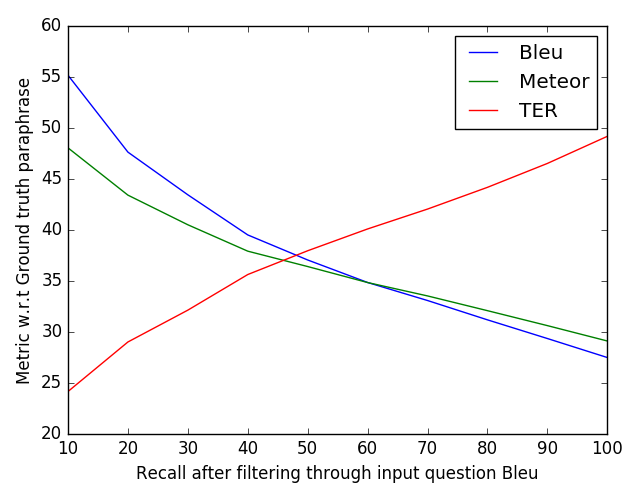}
    \end{center}
    \vspace{-0.3in}
    \caption{\small Recall vs BLEU/METEOR/TER after filtering the results through Q1-BLEU for Quora dataset}
    \label{fig:recall-bleu}
    \vspace{-0.5em}
\end{figure}

In Table~\ref{tab:mscoco-results}, we report the results for MSCOCO dataset. For this dataset, we compare the results of our approach with existing approaches. As we can see, we have a significant improvement w.r.t.the baselines. Both variations of our supervised model i.e., VAE-SVG and VAE-SVG-eq perform better than the state-of-the-art with VAE-SVG performing slightly better than VAE-SVG-eq. We also evaluate with respect to best variant. The best variant is computed using different metrics, that is, BLEU and METEOR, however the best variant is not always guaranteed to perform better than average since best variant is computed with respect to the input question not based on the ground truth. When using the best variant, we get improvement in all three metrics in the case of non-beam search, however when experimented with generating paraphrases through beam-search, we get further improvement for METEOR and TER however these improvement are not as significant as for the Quora dataset, as you will see below. This could be because MSCOCO is an image captioning dataset which means that dataset does not contain fully formed grammatical sentences, as one can see from the examples in Table~\ref{tab:good-examples-mscoco}. In such cases, beam search is not able to capture the structure of the sentence construction. When comparing our results with the state-of-the-art baseline, the average metric of the VAE-SVG model is able to give a 10\% absolute point performance improvement for the TER metric, a significant number with respect to the difference between the best and second best baseline which only stands at 2\% absolute point. For the BLEU and METEOR, our best results are $4.7\%$ and $4\%$  absolute point improvement over the state-of-the-art.

In Table~\ref{tab:quora-results}, we report results for the Quora dataset. As we can see, both variations of our model perform significantly better than unsupervised VAE and VAE-S, which is not surprising. We also report the results on different training sizes, and as expected, as we increase the training data size, results improve. Comparing the results across different variants of supervised model, VAE-SVG-eq performs the best. This is  primarily due to the fact that in VAE-SVG-eq, the parameters of the input question encoder are shared by the encoding side and the decoding side. We also experimented with generating paraphrases through beam-search, and, unlike MSCOCO, it turns out that beam search improves the results significantly. This is primarily because beam-search is able to filter out the paraphrases which had only few common terms with the input question. When comparing the best variant of our model with unsupervised model (VAE), we are able to get more than 27\% absolute point (more than 3 times) boost in BLEU score, and more than 19\% absolute point (more than 2 times) boost in METEOR; and when comparing with VAE-S, we are able to get a boost of almost 19\% absolute points in BLEU (2 times) and more than 10\% absolute points in METEOR (1.5 times).
\vspace{-0.2in}
\begin{table}[h]
\centering
\scriptsize
\caption{\scriptsize Our Human evaluation results for paraphrase generation.}
\label{tab:humanevaluation}
\begin{tabular}{|l|l|l|l|}
\hline
\textbf{Dataset} & \textbf{Input} & \textbf{Relevance} & \textbf{Readability} \\ \hline
MSCOCO   &    Ground Truth & 3.38 & 4.68  \\ \hline
  &   System Output & 3.0 & 3.84   \\       \hline
Quora   &    Ground Truth & 4.82 & 4.94  \\ \hline
  &   System Output & 3.57 & 4.08   \\       \hline
\end{tabular}
\end{table}

The results of the {\bf qualitative human evaluation} are shown in Table \ref{tab:humanevaluation}. From the Table, we see that our method produces results which are close to the ground truth for both metrics \textit{Readability} and \textit{Relevance}. Note that \textit{Relevance} of the MSCOCO dataset is $3.38$ which is far from a perfect score of 5 because unlike Quora, MSCOCO dataset is an image caption dataset, and therefore allows for a larger variation in the human annotations.

Note that one can use the metric between the variant and the input question to provide filtering in the case of multiple variants, or even to decide if a variant needs to be reported or not. So in order to make the system more practical (a high precision system), we choose to report the variant only when the confidence in the variant is more than a threshold. We use the metric between \textit{input question} and the variant to compute this confidence. Naturally this thresholding reduces the recall of the system. In Figure~\ref{fig:recall-bleu}, we plot the recall for Quora dataset, after thresholding the confidence (computed using the BLEU between the variant and the input question), and the average metrics for those candidates that pass the threshold. Interestingly, we can increase the BLEU score of the system as much as up to $55\%$ at the recall of $10\%$. Plots generated using other metrics such as METEOR and TER showed a similar trend.

\section{Conclusion}
In this paper we have proposed a deep generative framework, in particular, a Variational Autoencoders based architecture, augmented with sequence-to-sequence models, for generating paraphrases. Unlike traditional VAE and unconditional sentence generation model, our model conditions the encoder and decoder sides of the VAE on the input sentence, and therefore can generate \textit{multiple} paraphrases for a given sentence in a principled way. We evaluate the proposed method on a general paraphrase generation dataset, and show that it outperforms the state-of-the-art by a significant margin, without any hyper-parameter tuning. 
We also evaluate our approach on a recently released question paraphrase dataset, and demonstrate its remarkable performance. The generated paraphrases are not just semantically similar to the original input sentence, but also able to capture new concepts related to the original sentence. 
%Another remarkable aspect of our framework is its simplicity and modular architecture. Although the current architecture uses a basic VAE and LSTM modules, these can be replaced with more sophisticated building blocks (e.g., LSTMs replaced by GRUs).
\small
\bibliographystyle{aaai}
\bibliography{query_sum_ref}

\begin{thebibliography}{}

\bibitem[\protect\citeauthoryear{Bahdanau, Cho, and
  Bengio}{2014}]{bahdanau2014neural}
Bahdanau, D.; Cho, K.; and Bengio, Y.
\newblock 2014.
\newblock Neural machine translation by jointly learning to align and
  translate.
\newblock {\em arXiv preprint arXiv:1409.0473}.

\bibitem[\protect\citeauthoryear{Bowman \bgroup et al\mbox.\egroup
  }{2015}]{bowman2015generating}
Bowman, S.~R.; Vilnis, L.; Vinyals, O.; Dai, A.~M.; Jozefowicz, R.; and Bengio,
  S.
\newblock 2015.
\newblock Generating sentences from a continuous space.
\newblock {\em arXiv preprint arXiv:1511.06349}.

\bibitem[\protect\citeauthoryear{Chung \bgroup et al\mbox.\egroup
  }{2015}]{chung2015recurrent}
Chung, J.; Kastner, K.; Dinh, L.; Goel, K.; Courville, A.~C.; and Bengio, Y.
\newblock 2015.
\newblock A recurrent latent variable model for sequential data.
\newblock In {\em Advances in neural information processing systems},
  2980--2988.

\bibitem[\protect\citeauthoryear{Fader, Zettlemoyer, and
  Etzioni}{2014}]{fader2014open}
Fader, A.; Zettlemoyer, L.; and Etzioni, O.
\newblock 2014.
\newblock Open question answering over curated and extracted knowledge bases.
\newblock In {\em Proceedings of the 20th ACM SIGKDD international conference
  on Knowledge discovery and data mining},  1156--1165.
\newblock ACM.

\bibitem[\protect\citeauthoryear{Figueroa and
  Neumann}{2013}]{figueroa2013learning}
Figueroa, A., and Neumann, G.
\newblock 2013.
\newblock Learning to rank effective paraphrases from query logs for community
  question answering.
\newblock In {\em AAAI}, volume~13,  1099--1105.

\bibitem[\protect\citeauthoryear{Goodfellow, Bengio, and
  Courville}{2016}]{goodfellow2016deep}
Goodfellow, I.; Bengio, Y.; and Courville, A.
\newblock 2016.
\newblock {\em Deep learning}.
\newblock The MIT Press.

\bibitem[\protect\citeauthoryear{Graves, Jaitly, and
  Mohamed}{2013}]{graves2013hybrid}
Graves, A.; Jaitly, N.; and Mohamed, A.-r.
\newblock 2013.
\newblock Hybrid speech recognition with deep bidirectional lstm.
\newblock In {\em Automatic Speech Recognition and Understanding (ASRU), 2013
  IEEE Workshop on},  273--278.
\newblock IEEE.

\bibitem[\protect\citeauthoryear{Hassan \bgroup et al\mbox.\egroup
  }{2007}]{hassan2007unt}
Hassan, S.; Csomai, A.; Banea, C.; Sinha, R.; and Mihalcea, R.
\newblock 2007.
\newblock Unt: Subfinder: Combining knowledge sources for automatic lexical
  substitution.
\newblock In {\em Proceedings of the 4th International Workshop on Semantic
  Evaluations},  410--413.
\newblock Association for Computational Linguistics.

\bibitem[\protect\citeauthoryear{Hochreiter and
  Schmidhuber}{1997}]{hochreiter1997long}
Hochreiter, S., and Schmidhuber, J.
\newblock 1997.
\newblock Long short-term memory.
\newblock {\em Neural computation} 9(8):1735--1780.

\bibitem[\protect\citeauthoryear{Hu \bgroup et al\mbox.\egroup
  }{2017}]{hu2017controllable}
Hu, Z.; Yang, Z.; Liang, X.; Salakhutdinov, R.; and Xing, E.~P.
\newblock 2017.
\newblock Controllable text generation.
\newblock {\em arXiv preprint arXiv:1703.00955}.

\bibitem[\protect\citeauthoryear{Kingma and Welling}{2013}]{kingma2013auto}
Kingma, D.~P., and Welling, M.
\newblock 2013.
\newblock Auto-encoding variational bayes.
\newblock {\em arXiv preprint arXiv:1312.6114}.

\bibitem[\protect\citeauthoryear{Kingma and Welling}{2014}]{kingma2014auto}
Kingma, D.~P., and Welling, M.
\newblock 2014.
\newblock Auto-encoding variational bayes.
\newblock In {\em ICLR}.

\bibitem[\protect\citeauthoryear{Kingma \bgroup et al\mbox.\egroup
  }{2014}]{kingma2014semi}
Kingma, D.~P.; Mohamed, S.; Rezende, D.~J.; and Welling, M.
\newblock 2014.
\newblock Semi-supervised learning with deep generative models.
\newblock In {\em Advances in Neural Information Processing Systems},
  3581--3589.

\bibitem[\protect\citeauthoryear{Kozlowski, McCoy, and
  Vijay-Shanker}{2003}]{kozlowski2003generation}
Kozlowski, R.; McCoy, K.~F.; and Vijay-Shanker, K.
\newblock 2003.
\newblock Generation of single-sentence paraphrases from predicate/argument
  structure using lexico-grammatical resources.
\newblock In {\em Proceedings of the second international workshop on
  Paraphrasing-Volume 16},  1--8.
\newblock Association for Computational Linguistics.

\bibitem[\protect\citeauthoryear{Lavie and Agarwal}{2007}]{lavie2007meteor}
Lavie, A., and Agarwal, A.
\newblock 2007.
\newblock Meteor: An automatic metric for mt evaluation with high levels of
  correlation with human judgments.
\newblock In {\em Proceedings of the Second Workshop on Statistical Machine
  Translation},  228--231.
\newblock Association for Computational Linguistics.

\bibitem[\protect\citeauthoryear{Lin \bgroup et al\mbox.\egroup
  }{2014}]{lin2014microsoft}
Lin, T.-Y.; Maire, M.; Belongie, S.; Bourdev, L.; Girshick, R.; Hays, J.;
  Perona, P.; Ramanan, D.; Zitnick, C.~L.; and Dollar, P.
\newblock 2014.
\newblock Microsoft coco: Common objects in context.
\newblock {\em arXiv preprint arXiv:1405.0312}.

\bibitem[\protect\citeauthoryear{Madnani and
  Dorr}{2010}]{madnani2010generating}
Madnani, N., and Dorr, B.~J.
\newblock 2010.
\newblock Generating phrasal and sentential paraphrases: A survey of
  data-driven methods.
\newblock {\em Computational Linguistics} 36(3):341--387.

\bibitem[\protect\citeauthoryear{Madnani, Tetreault, and
  Chodorow}{2012}]{madnani2012re}
Madnani, N.; Tetreault, J.; and Chodorow, M.
\newblock 2012.
\newblock Re-examining machine translation metrics for paraphrase
  identification.
\newblock In {\em Proceedings of the 2012 Conference of the North American
  Chapter of the Association for Computational Linguistics: Human Language
  Technologies},  182--190.
\newblock Association for Computational Linguistics.

\bibitem[\protect\citeauthoryear{McKeown}{1983a}]{McKeown:1983:PQU:973241.973242}
McKeown, K.~R.
\newblock 1983a.
\newblock Paraphrasing questions using given and new information.
\newblock {\em Comput. Linguist.} 9(1):1--10.

\bibitem[\protect\citeauthoryear{McKeown}{1983b}]{mckeown1983paraphrasing}
McKeown, K.~R.
\newblock 1983b.
\newblock Paraphrasing questions using given and new information.
\newblock {\em Computational Linguistics} 9(1):1--10.

\bibitem[\protect\citeauthoryear{Papineni \bgroup et al\mbox.\egroup
  }{2002}]{papineni2002BLEU}
Papineni, K.; Roukos, S.; Ward, T.; and Zhu, W.-J.
\newblock 2002.
\newblock Bleu: a method for automatic evaluation of machine translation.
\newblock In {\em Proceedings of the 40th annual meeting on association for
  computational linguistics},  311--318.
\newblock Association for Computational Linguistics.

\bibitem[\protect\citeauthoryear{Prakash \bgroup et al\mbox.\egroup
  }{2016}]{prakash2016neural}
Prakash, A.; Hasan, S.~A.; Lee, K.; Datla, V.; Qadir, A.; Liu, J.; and Farri,
  O.
\newblock 2016.
\newblock Neural paraphrase generation with stacked residual lstm networks.
\newblock {\em arXiv preprint arXiv:1610.03098}.

\bibitem[\protect\citeauthoryear{Quirk, Brockett, and
  Dolan}{2004}]{quirk2004monolingual}
Quirk, C.; Brockett, C.; and Dolan, W.~B.
\newblock 2004.
\newblock Monolingual machine translation for paraphrase generation.
\newblock In {\em EMNLP},  142--149.

\bibitem[\protect\citeauthoryear{Rezende, Mohamed, and
  Wierstra}{2014}]{rezende2014stochastic}
Rezende, D.~J.; Mohamed, S.; and Wierstra, D.
\newblock 2014.
\newblock Stochastic backpropagation and approximate inference in deep
  generative models.
\newblock {\em arXiv preprint arXiv:1401.4082}.

\bibitem[\protect\citeauthoryear{Semeniuta, Severyn, and
  Barth}{2017}]{semeniuta2017hybrid}
Semeniuta, S.; Severyn, A.; and Barth, E.
\newblock 2017.
\newblock A hybrid convolutional variational autoencoder for text generation.
\newblock {\em arXiv preprint arXiv:1702.02390}.

\bibitem[\protect\citeauthoryear{Snover \bgroup et al\mbox.\egroup
  }{2006}]{snover2006study}
Snover, M.; Dorr, B.; Schwartz, R.; Micciulla, L.; and Makhoul, J.
\newblock 2006.
\newblock A study of translation edit rate with targeted human annotation.
\newblock In {\em Proceedings of association for machine translation in the
  Americas}, volume 200.

\bibitem[\protect\citeauthoryear{Sohn, Lee, and Yan}{2015}]{sohn2015learning}
Sohn, K.; Lee, H.; and Yan, X.
\newblock 2015.
\newblock Learning structured output representation using deep conditional
  generative models.
\newblock In {\em Advances in Neural Information Processing Systems},
  3483--3491.

\bibitem[\protect\citeauthoryear{Sutskever, Vinyals, and
  Le}{2014}]{sutskever2014sequence}
Sutskever, I.; Vinyals, O.; and Le, Q.~V.
\newblock 2014.
\newblock Sequence to sequence learning with neural networks.
\newblock In {\em Advances in neural information processing systems},
  3104--3112.

\bibitem[\protect\citeauthoryear{Vinyals \bgroup et al\mbox.\egroup
  }{2015}]{vinyals2015show}
Vinyals, O.; Toshev, A.; Bengio, S.; and Erhan, D.
\newblock 2015.
\newblock Show and tell: A neural image caption generator.
\newblock In {\em Proceedings of the IEEE conference on computer vision and
  pattern recognition},  3156--3164.

\bibitem[\protect\citeauthoryear{Wubben, Van Den~Bosch, and
  Krahmer}{2010}]{wubben2010paraphrase}
Wubben, S.; Van Den~Bosch, A.; and Krahmer, E.
\newblock 2010.
\newblock Paraphrase generation as monolingual translation: Data and
  evaluation.
\newblock In {\em Proceedings of the 6th International Natural Language
  Generation Conference},  203--207.
\newblock Association for Computational Linguistics.

\bibitem[\protect\citeauthoryear{Yin \bgroup et al\mbox.\egroup
  }{2015}]{yin2015answering}
Yin, P.; Duan, N.; Kao, B.; Bao, J.; and Zhou, M.
\newblock 2015.
\newblock Answering questions with complex semantic constraints on open
  knowledge bases.
\newblock In {\em Proceedings of the 24th ACM International on Conference on
  Information and Knowledge Management},  1301--1310.
\newblock ACM.

\bibitem[\protect\citeauthoryear{Zhao \bgroup et al\mbox.\egroup
  }{2008}]{zhao2008combining}
Zhao, S.; Niu, C.; Zhou, M.; Liu, T.; and Li, S.
\newblock 2008.
\newblock Combining multiple resources to improve smt-based paraphrasing model.
\newblock In {\em ACL},  1021--1029.

\bibitem[\protect\citeauthoryear{Zhao \bgroup et al\mbox.\egroup
  }{2009}]{zhao2009application}
Zhao, S.; Lan, X.; Liu, T.; and Li, S.
\newblock 2009.
\newblock Application-driven statistical paraphrase generation.
\newblock In {\em Proceedings of the Joint Conference of the 47th Annual
  Meeting of the ACL and the 4th International Joint Conference on Natural
  Language Processing of the AFNLP: Volume 2-Volume 2},  834--842.
\newblock Association for Computational Linguistics.

\end{thebibliography}

\end{document}